\documentclass[10pt,twocolumn,letterpaper]{article}

\usepackage{cvpr}
\usepackage{times}
\usepackage{epsfig}
\usepackage{graphicx}
\usepackage{amsmath}
\usepackage{amssymb}

\usepackage{graphicx}
\usepackage{color}
\usepackage{amsmath}
\usepackage{algorithm}
\usepackage{algorithmic}
\usepackage{tabulary}
\usepackage{subfig}
\usepackage{booktabs}

\usepackage[pagebackref=true,breaklinks=true,letterpaper=true,colorlinks,bookmarks=false]{hyperref}

\cvprfinalcopy % *** Uncomment this line for the final submission

\def\cvprPaperID{2140} % *** Enter the CVPR Paper ID here

\ifcvprfinal\fi
\begin{document}

%%%%%%%%% TITLE
\title{Learning to Learn from Noisy Web Videos}

\author{
  Serena Yeung \\
  Stanford University\\
  {\tt\small serena@cs.stanford.edu} \\
  \and
  Vignesh Ramanathan \\
  Stanford University \\
  {\tt\small vigneshr@stanford.edu} \\
  \and
  Olga Russakovsky \\
  Carnegie Mellon University \\
  {\tt\small olgarus@cmu.edu} \\
  \and
  Liyue Shen \\
  Stanford University \\
  {\tt\small liyues@stanford.edu} \\
  \and
  Greg Mori \\
  Simon Fraser University \\
  {\tt\small mori@cs.sfu.ca} \\
  \and
  Li Fei-Fei \\
  Stanford University \\
  {\tt\small feifeili@cs.stanford.edu} \\
}

\maketitle
%\thispagestyle{empty}

%%%%%%%%% ABSTRACT
\begin{abstract}

Understanding the simultaneously very diverse and intricately fine-grained set of possible human actions is a critical open problem in computer vision. Manually labeling training videos is feasible for some action classes but doesn't scale to the full long-tailed distribution of actions. A promising way to address this is to leverage noisy data from web queries to learn new actions, using semi-supervised or ``webly-supervised'' approaches. However, these methods typically do not learn domain-specific knowledge, or rely on iterative hand-tuned data labeling policies. In this work, we instead propose a reinforcement learning-based formulation for selecting the right examples for training a classifier from noisy web search results. Our method uses Q-learning to learn a data labeling policy on a small labeled training dataset, and then uses this to automatically label noisy web data for new visual concepts. Experiments on the challenging Sports-1M action recognition benchmark
as well as on additional fine-grained and newly emerging action classes demonstrate that our method is able to learn good labeling policies for noisy data and use this to learn accurate visual concept classifiers.

\end{abstract}

%%%%%%%%% BODY TEXT
\section{Introduction}

Humans are a central part of many visual scenes, and understanding human actions in videos is an important problem in computer vision. However, a key challenge in action recognition is scaling to the long tail of actions. In many practical applications, we would like to quickly and cheaply learn classifiers for new target actions where annotations are scarce, e.g. fine-grained, rare or niche classes. Manually annotating data for every new action becomes impossible, so there is a need for methods that can automatically learn from readily available albeit noisy data sources.

\begin{figure}[t]
\begin{center}
\includegraphics[width=\linewidth]{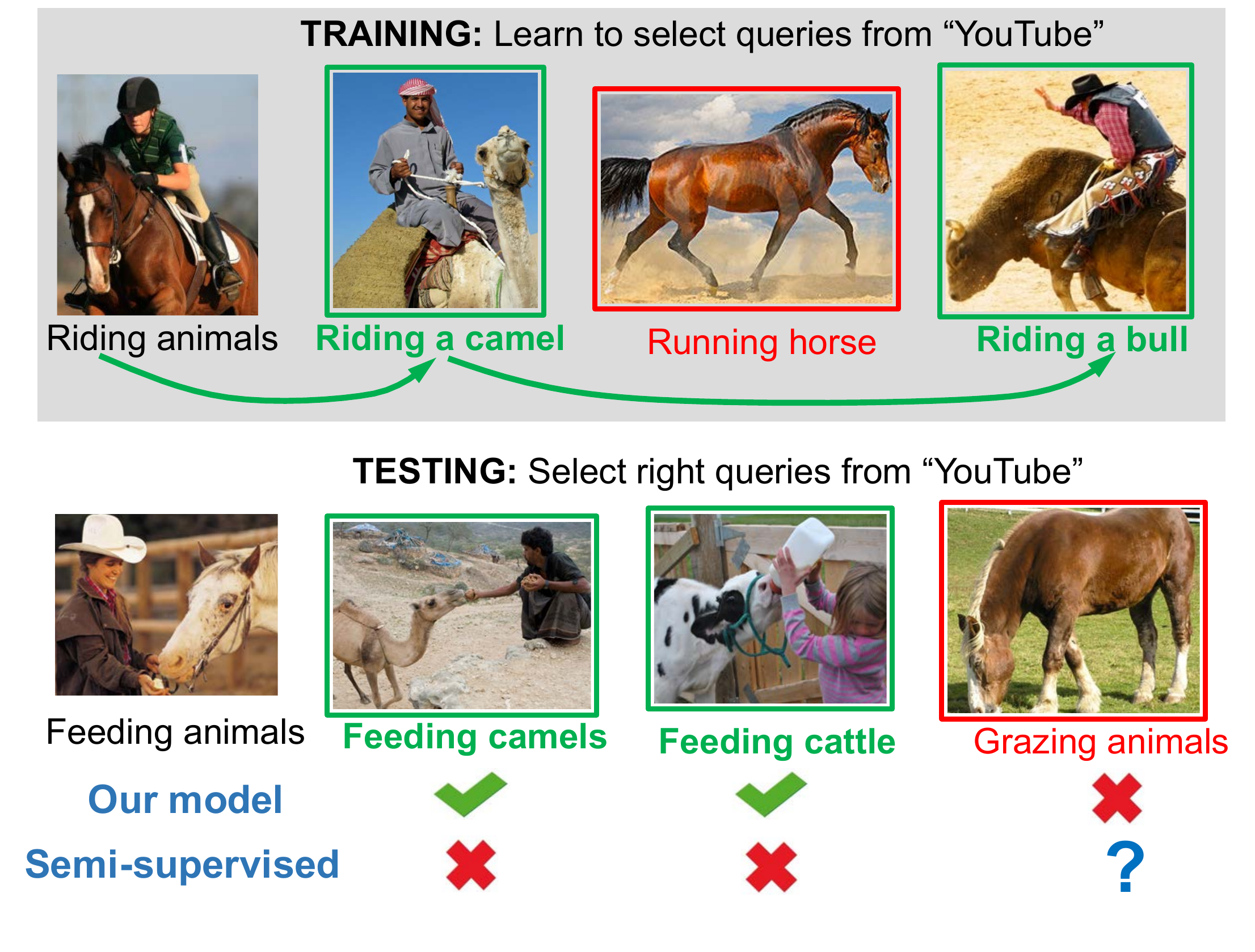}
\end{center}
   \caption{Our model uses a set of annotated data to learn a policy for how to label data for new, unseen classes. This enables learning domain-specific knowledge and how to select diverse exemplars while avoiding semantic drift. For example, it can learn from training data that human motion cues are important for actions involving animals (e.g. ``riding animals'') while animal appearance is not. This knowledge can be applied at test time to label noisy data for new classes such as ``feeding animals'', while traditional semi-supervised methods would label based on visual similarity.}
\label{fig:pull}
\label{fig:onecol}
\vspace{-12pt}
\end{figure}

%Humans are a central part of many visual scenes, and understanding human actions is an important problem in computer vision. However, a key challenge in action recognition is scaling to the long tail of action categories. In many practical applications, we would like to cheaply learn classifiers for new categories of interest, which may be fine-grained or niche concepts. Manual data labeling quickly becomes prohibitively expensive, especially in videos. As a results, there is a need for methods that can automatically acquire labeled data for new concepts from noisy sources, without traditional manual labeling.

A promising approach is to leverage noisy data from web queries. Training models for new classes using the data returned by web queries has been proposed as an alternative to expensive manual annotation~\cite{chen2013neil,LEVAN,li2010optimol,schroff2011harvesting}. Methods for automated labeling of new classes include traditional semi-supervised learning approaches~\cite{Joachims_ICML99,Zhou_NIPS04,Zhu_ICML03} as well as webly-supervised approaches~\cite{chen2013neil,LEVAN,li2010optimol}. However, these methods typically rely on iterative hand-tuned data labeling policies. This makes it difficult to dynamically manage the risk trade-off between exemplar diversity and semantic drift. Going further, as a result these methods typically cannot learn domain-specific knowledge. For example, when learning an action recognition model from a set of videos returned by YouTube queries, videos prominently featuring humans are more likely to be positives while those without are more likely to be noise; this intuition is difficult to manually quantify and encode. Even more, when learning an animal-related action such as ``feeding animals'', videos containing the action with different animals are likely to be useful positives even though their visual appearance may be  different (Fig.~\ref{fig:pull}). Such diverse class-conditional data selection policies are impossible to manually encode. This intuition inspires our work on learning data selection policies for noisy web search results.

Our key insight is that good data labeling policies can be learned from existing manually annotated datasets. Intuitively, a good policy labels noisy data in a way where a classifier trained on the labels would achieve high classification accuracy on a manually annotated held-out set.
%in a way that enables training a classifier with these labels to achieves high accuracy on a manually annotated dataset, i.e., 
%that labels noisy data in a way where a classifier trained on the data achieves strong classification performance on a manually annotated dataset should also be a good policy for labeling new classes. 
%In other words, the policy should be learned to maximize accuracy of a downstream classifier.
Although data labeling is a non-differentiable action, this can be naturally achieved in a reinforcement learning setting, where actions correspond to labeling of examples and the reward is the effect on downstream classifier accuracy.

Concretely, we introduce a joint formulation of a Q-learning agent~\cite{watkins1992q} and a class recognition model. In contrast to related webly-supervised approaches~\cite{chen2013neil,li2010optimol}, the data collection and classifier training steps are not disjoint but rather integrated into a single unified framework. The agent selects web search examples to label as positives, which are then used to train the recognition model. A significant challenge is the choice of the state representation, and we introduce a novel representation based on the distribution of classifier scores output by the recognition model. At training time, the model uses a dataset of labeled training classes to learn a data labeling policy, and at test time the model can use this policy to label noisy web data for new unseen classes.

In summary, our main contribution is a principled formulation for learning how to label noisy web data, using a reinforcement learning framework. To enable this, we also introduce a novel state representation in terms of the classifier score distributions from a jointly trained recognition model. We demonstrate our approach first in the controlled setting of MNIST, then on the large-scale Sports-1M video benchmark~\cite{Karpathy_CVPR14}. Finally, we show that our method can be used for labeling newly emerging and fine-grained categories where annotated data is scarce.

\section{Related work}

%\todo{Olga: (a) I believe this is not online learning, (b) I moved it to the end since I think it broke flow as section 2 but maybe that's not good, and  (c) Need to take a pass and clean up, hopefully Greg?}

%online learning~\cite{widmer1996learning} methods like

The difficulty of building large-scale annotated datasets has inspired methods such as~\cite{Chen15,chen2013neil,chen2015webly,gan2016webly,li2010optimol,LEVAN,schroff2011harvesting,liang2016learning} which attempts to learn visual (or text~\cite{carlson2010toward}) models from noisy web-search results. Such methods usually focus on iteratively gathering examples and using them to improve the visual classifier. Often specific constraints or hand-tuned rules are used for  data collection, and successive iterations can cause the model to deviate from the initial concept. We overcome these limitations by automatically learning robust data collection policies resulting in accurate classifiers.

%our learned data collection policies are robust to such semantic drift

%Often, specific constraints or assumptions about relationship between labels are used to restrict the noise added by web-search. Also, . However, policies learned on labelled data collections could be robust to such semantic-drifts and do not require prior knowledge about the domain.

The task of semi-supervised learning also works with limited annotated examples.
Popular approaches like transductive SVM \cite{Joachims_ICML99}, label
spreading~\cite{Zhou_NIPS04} and label propagation~\cite{Zhu_ICML03} induce labels for
unannotated examples to explain their distribution. Recent approaches~\cite{Kingma_NIPS14,li2015semi,Pitelis_KDD14,ranzato2008semi,Rifai_NIPS11,weston2012deep} learn an embedding space which captures this distribution. However,
these methods do not learn domain-specific knowledge which can help in
pruning noisy examples or understanding the multiple subcategories
within a class. In contrast, our learned policies adjust for such biases
in web-examples of a specific domain.

Approaches like co-segmentation \cite{icoseg,joulin2010discriminative}, multiple instance learning~\cite{Andrews_NIPS02} and zero-shot learning~\cite{frome2013devise,palatucci2009zero,rohrbach2011evaluating,xu2016multi} do incorporate domain-specific knowledge. However, unlike our method they do not utilize the large amount of
web-search data available for test classes.

%Weakly supervised methods like
% and Multiple Instance
%%Learning jointly segment examples belonging to multiple
%classes.  Zero-shot learnin
%attempts to learn domain-specific embedding only using labelled examples during
%training.  While these approaches incorporate
%domain-knowledge,

Our setup is similar in spirit to meta-learning~\cite{brazdil2008metalearning,feurer2015efficient}, which attempts to identify both the correct learning algorithm and the parameters required for high accuracy. However, we target a unique goal of building a
good dataset for a given class from a large set of noisy web-search examples.

Our key insight is that we can learn policies to directly optimize our goal of choosing a positive set (nondifferentiable actions) leading to an accurate visual classifier, by formulating it in a reinforcement learning setup.  We leverage recent advances that enable the use of deep neural networks as function approximators in deep Q-learning~\cite{mnih2015human}, and which has shown successful performance in learning policies for game-playing~\cite{mnih2015human}, large-scale control problems~\cite{dulac2015reinforcement}, and simple algorithms~\cite{zaremba2015learning}.

%Specifically, we learn policies for how to do this in a Q-learning~\cite{watkins1992q} formulation.

\section{Method}

\begin{figure*}
  \centering
  \includegraphics[width=0.9\linewidth]{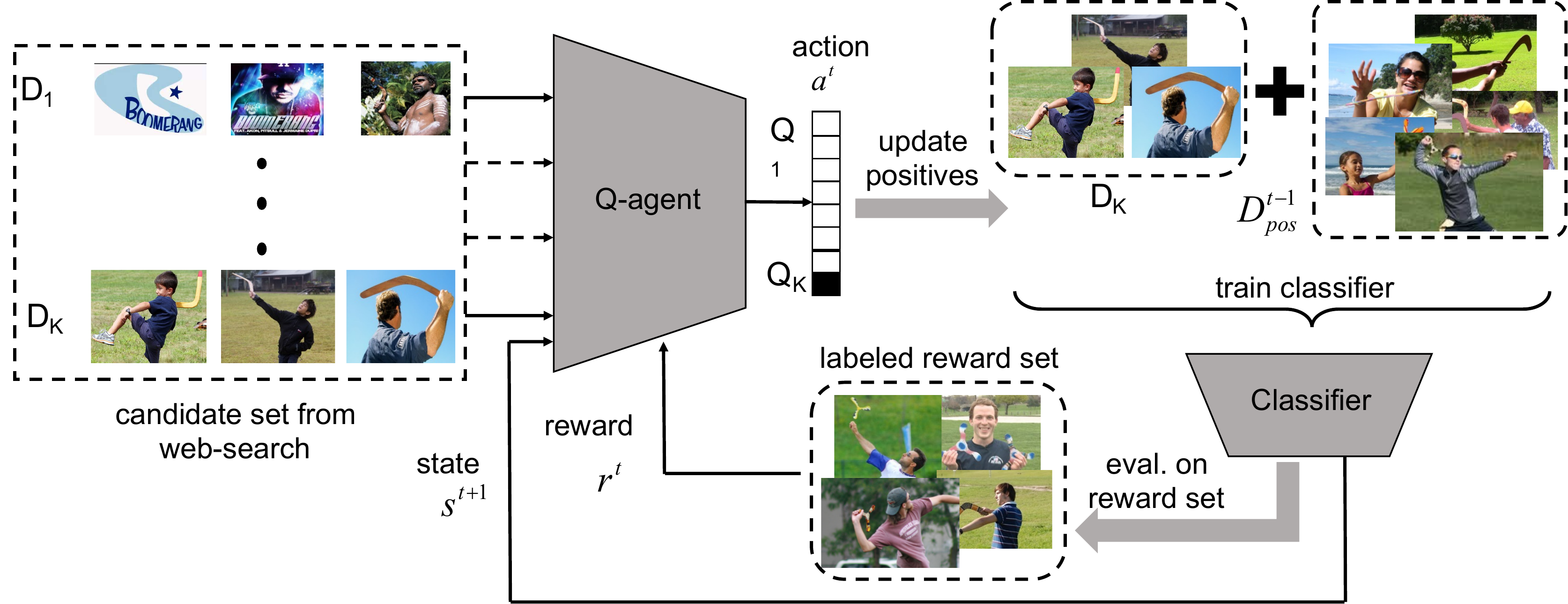}
  \caption{\small{Overview of our model. We learn a classifier for a given visual concept using a candidate set of examples obtained from web search. At each time step $t$ we use the Q-learning agent to select examples, e.g., $D_K$, to add to our existing set of positive  examples $D_{pos}^{t-1}$. The examples are then used to train a visual classifier. The classifier both updates the agent's state $s^{t+1}$ and provides a reward $r^{t}$. At test time the trained agent can be used to automatically select positive examples from web search results for any new visual concept.}
  }
  \vspace{-12pt}
  \label{fig:overview}
\end{figure*}

The goal of our method is to automatically learn an accurate classifier of a visual concept directly from noisy web search results. We refer to these noisy search results as the \textbf{candidate set} $D_{cand} = \{D_1, \dots, D_K\}$, and wish to select a good subset of positives from $D_{cand}$. Such weakly supervised data typically contains diverse subclasses, and the key challenge is to capture this  diversity without succumbing to semantic drift. These properties are difficult to objectively quantify. Hence, we want a model which can learn them from existing labeled datasets. One way to achieve this is through an iterative strategy for positive selection, where the model is aware of the positives chosen so far, so that it can promote diversity in future selections. On the other hand, it also needs to avoid semantic drift by being aware of the remaining candidates and learning to estimate long-term change in classifier accuracy. This can be elegantly achieved through a Q-learning formulation.

Hence, our model consists of two components as shown in Fig.~\ref{fig:overview}: (1) a \emph{classifier model} trained using selected positive examples from the candidate set, and (2) a \emph{Q-learning agent} which repeatedly selects the positives from the candidate set to train the classifier model.

Note that the visual classes used for training and testing our method are disjoint. For every visual class used during training, in addition to the {\bf candidate set}, we are also provided a {\bf reward set}, which is a set of examples  annotated with the presence or absence of the target class. The reward set is used to evaluate our model's ability to produce a good classifier from the candidate set. At test time,  we are given just the {\bf candidate set} for new visual classes.

%Each subset $D_k$ of our candidate set contains one or more visual examples; concretely, each $D_k$ corresponds to one page of results returned by one search query related to the target concept.

\subsection{Classifier model}
\label{sec:classifier_model}

The classifier model corresponds to the current visual class considered during a data collection
\emph{episode} and is trained simultaneously with the agent's data collection policy.
At the beginning of an episode, the classifier is seeded
with a small set of $S$ positive examples $D_{seed}=\{x_1, ..., x_S\}$, as well
as a set of negative examples $D_{neg}=\{x_1, ..., x_N\}$. In the case of a
video search-engine, we assume that the top few retrieved examples are of high enough
quality to serve as the seed positives. A random collection of videos from
multiple unrelated searches are used to construct the negative set. At each time step $t$, the Q-learning agent makes a selection $a_t$ corresponding to examples
$D_{a_t}$ to be removed from the candidate set $D_{cand}$ and added to the positive training set. The classifier is then trained to distinguish between $D_{pos} = \{D_{seed} \cup D_{a_1} \cup ... \cup D_{a_t} \}$ and $D_{neg}$.
The classifier is treated as a black box by the agent; in our experiments, we use a multi-layer perceptron.

%\noindent {\bf Candidates.} We are given an unlabeled candidate set $D_{cand} = \{Q_1, \dots, Q_K\}$ of noisy web search results for a visual class. The agent is tasked with automatically selecting positive examples for training a visual classifier. Such weakly supervised data typically contains a diversity of subclasses, and the key challenge is to learn models which can capture this  diversity without succumbing to semantic drift. Each subset $Q_k$ of our candidate set contains one or more visual examples; concretely, each $Q_k$ corresponds to one page of results returned by one search query related to the target concept. Operating over subsets allows the model to reason at a subclass level, efficiently selecting the subclassses  useful for improving the visual classifier.

%\noindent {\bf Iterative labeling.} 

%t timestep $t$, the
%recognition model is therefore updated using training data consisting of
%positive set
%\begin{equation}

%\end{equation} and negative set $D_{neg}$.  The recognition model is treated as a black-box
%classifier by the agent; in our experiments, we use a
%two-layer perceptron.

\subsection{Q-learning agent}

The core of our model is a Q-learning agent.  Each episode observed by the agent corresponds to data collection for a specific class.  At each timestep $t$, the
agent observes the current state $s_t$, and selects an action $a_t$ from a
discrete set of actions $\mathcal{A}=\{1,...,K\}$. The action updates the
classifier as in Sec.~\ref{sec:classifier_model}, and the agent receives a reward $r_t$ and the next state
observation $s_{t+1}$. The agent's goal at each timestep is to choose the
action that maximizes the future discounted reward $R_t = \sum_{t'=t}^T\gamma^{t'-t}r_{t'}$,
 where an episode terminates at
time $T$ and $\gamma$ is the discount factor.

There are several key decisions: (1) How to encode the current \emph{state} of the agent? (2) How to translate this state to a Q-value which can inform the agent's action? (3) How to formulate a reward function that incentivizes the agent to select optimal examples from the candidate set?

%How to formulate a reward that compares the quality of chosen examples to labelled examples?  2. How to handle the search-results of multiple queries which yield relevant examples for a given concept?  3.

%Selecting $a_t$ adds candidate subset $Q_{a_t}$ to the the positive set $D_{pos}$ used to train the recognition model.

\noindent \textbf{State representation.}
%Formulating an appropriate state representation $s$ is crucial in enabling the
%agent to learn good policies.
Our insight in formulating the state representation is that in order to improve the
visual classifier, the best examples to use may not be the ones that are the strongest positives according to the current classifier. Instead, it may be better to add some
examples with diversity that increase entropy in the positive set. However,
too much diversity will cause semantic drift. In order to reason about this,
the agent needs to fully understand the distribution of data in the previously selected
positive examples $D_{pos}$, in the negative set $D_{neg}$, and in the remaining noisy set $D_{cand}$.

\begin{figure}
    \centering
    \footnotesize
  \begin{tabular}{c|c}
    Generally correct and similar & Generally correct and dissimilar \\
    \includegraphics[width=0.45\linewidth]{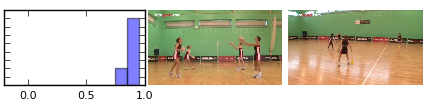} &
    \includegraphics[width=0.45\linewidth]{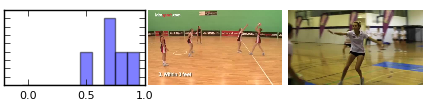} \\
    \hline
    Generally incorrect and similar & Generally incorrect and dissimilar\\
    \includegraphics[width=0.45\linewidth]{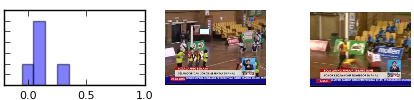}&
    \includegraphics[width=0.45\linewidth]{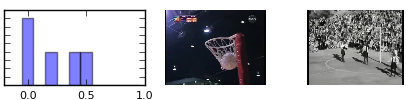} \\
    \end{tabular}
    \caption{\small{Histograms of classifiers scores for several query buckets with different characteristics, for an example class ``Netball". Example videos from each query are also shown. Buckets with correct vs. incorrect videos have high vs. low scores, while appearance diversity is reflected in histogram peakiness.}}
    \label{fig:histogram_examples}
\end{figure}

We therefore formulate the agent's state using the distribution of the classifier scores. Concretely, the state representation is $s=\{H_{pos}, H_{neg},\{H_{D_1},...,H_{D_K}\}, P\}$ where $H_{pos}, H_{neg},\{H_{D_1},...,H_{D_K}\}$ are histograms of classifier
scores for the positive set, the negative set, and each candidate subset, respectively. $P$ is the proportion of desired number of positives already obtained. The histograms capture the diversity and  ``prototypicality'' of each set (Fig.~\ref{fig:histogram_examples}).

%in both currently labeled
%data, as well as the candidate set. In particular, it needs to understand the
%distribution of the data relative to the recognition model.  for the examples in
%as well as the quantity of data collected so far. The state
%representation $s$ is given by

%\begin{eqnarray}
%  ,
%\end{eqnarray} where where the histogram distribution captures both the
%class diversity and prototypicality within each set, and

\begin{figure}[t]
  \centering
  \includegraphics[width=0.9\linewidth]{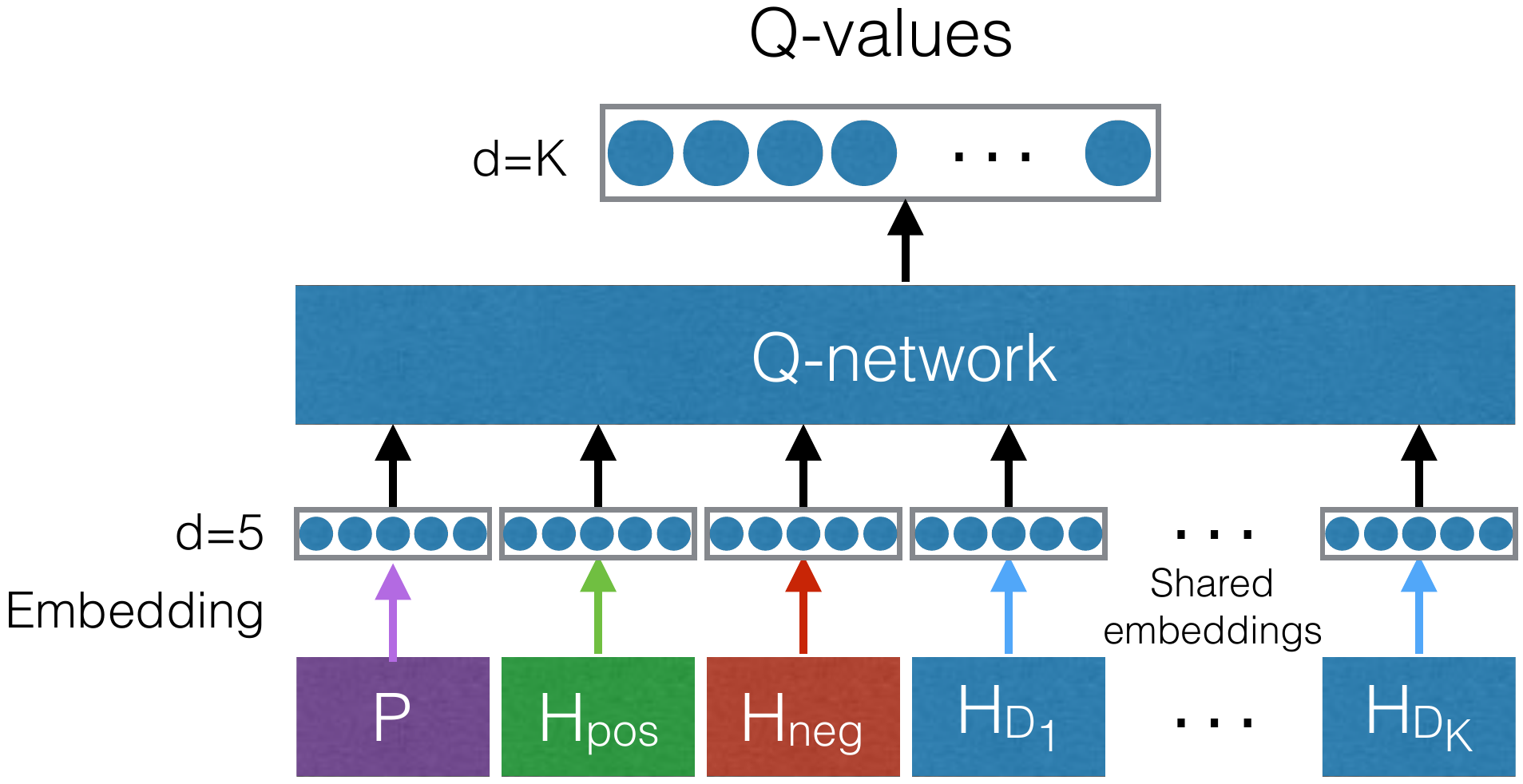}
  \caption{\small{The Q-network, which at each time-step chooses a subset of examples from $D_1, \dots, D_K$. The state representation is $s=\{H_{pos}, H_{neg},\{H_{D_1},...,H_{D_K}\}, P\}$ where $H_{pos}, H_{neg},\{H_{D_1},...,H_{D_K}\}$ are histograms of classifier scores for the positive set, the negative set, and each candidate subset. $P$ is the proportion of desired number of positives already obtained.} }
  \label{fig:qnet}
  %\vspace{-14pt}
\end{figure}

\noindent \textbf{Q-network.}
The agent takes an action $a_t$ at time $t$ using a policy $a_t = \max_{a}
Q(s_t, a)$, where $s_t$ is the state representation above. The
Q-value $Q(s_t, a)$ is determined by a neural network as illustrated in
Fig.~\ref{fig:qnet}.  Concretely, $\alpha_a = \phi \left(H_{pos}, H_{neg}, H_{D_a} ; \theta\right)$ where $\phi(.)$ is a multi-layer perceptron. We use $Q(s,a) = \text{softmax}(\alpha_a; \tau)$ where $\tau$ is a temperature parameter and helped performance in practice.

%The network assigns a Q-value to each of the $K$ candidate
%set options based on their classification score histograms, as well as the
%histograms of the negative and positive sets chosen so far.

%\begin{eqnarray}
%  \alpha_a & = & \phi \left(H_{pos}, H_{neg}, H_{Q_a} ; \theta\right) \\ \nonumber
%  Q(s, a) & = & ,
%\end{eqnarray}where $\tau$ is the temperature parameter of
%the softmax and .

%At any time $t$, if the  option $Q_k$ is selected by the Q-network,
%then the candidate set is updated to exclude this
%option at the next instant: $D_{\text{learn}}^{t+1} = D_{\text{learn}}^{t} \setminus Q_k$.

% The Q-network architecture, whose action at each time-step chooses one subset of examples from $D_1, \dots, D_K$. As shown, each subset is represented by the histogram of scores $H_{D_k}$ assigned by the classifier model. This is concatenated with the score histograms of the selected positive examples $H_{pos}$ and negative examples $H_{neg}$ from the candidate set. The concatenated histograms are fed to an MLP with shared weights for each subset, and scores assigned for each subset are normalized by a softmax.

\noindent \textbf{Reward function.}
The agent is incentivized to select the optimal examples from $D_{cand}$ for training a good classifier.
We capture this intuition by setting the reward at time $t$ to be the change in
the classifier's accuracy after updating its positive set with the newly chosen
examples $D_{a_t}$. Accuracy is computed on the held-out annotated data
$D_{reward}$. This reward is only available during training.

%examples which will eventually
%train a good classifier. Hence, the reward for an action chosen by the
%Q-network should reflect the subsequent change in the classifier's performance.

\subsection{Training and testing}
We train the agent using Q-learning~\cite{watkins1992q}, a standard reinforcement learning
algorithm that can be used to learn policies for an agent interacting with an
environment. In our case the environment is the visual classifier model. Each episode during training
corresponds to a specific visual class, where the agent selects
the positive examples of the class from a collection of web-search videos.

The Q-network parameters $\theta$ are learned by optimizing:
%\begin{eqnarray}
%  \small{L_i(\theta_i) & = & \mathbb{E}_{s,a}[ \left(\mathbb{E}_{s'}\left[r+\gamma \max_{a'} Q(s',a';\theta_{i-1}) | s,a \right]} \\ \nonumber
%  & & \small{- Q(s,a;\theta_i)\right) ^2 ],} \nonumber
%\end{eqnarray}

\begin{equation}
  \small{L_i(\theta_i)} =\small{\mathbb{E}_{s,a}\left[ \left( V(\theta_{i-1}) - Q(s,a;\theta_i) \right) ^2 \right],}
  \vspace{-2pt}
\end{equation}

where $i$ is an iteration of optimization and
\begin{equation}
    \small{V(\theta_{i-1})} = \small{\mathbb{E}_{s'}\left[r+\gamma \max_{a'} Q(s',a';\theta_{i-1}) | s,a \right].}
    \vspace{-2pt}
\end{equation}

We optimize it using stochastic gradient descent and
experience replay, with random minibatches of past experience $(s_t, a_t, r_t,
s_{t+1})$ sampled for training.

%\subsection{Testing}
The agent is trained to learn data collection policies which can generalize to
unseen visual classes.
%It is tested on new visual classes not encountered during training.
At test time, the agent and the classifier are again run simultaneously on $D_{cand}$ for a new class, but without access to any labeled examples $D_{reward}$. The agent selects videos using the learned greedy policy: $a_t = \max_a Q(s_t, a; \theta)$.

%Hence, the model is tested on a
%set of new classes which are different from the training classes. The Q-agent
%and the classifier are again run simultaneously for a new test class. During
%testing, the agent only sees the videos returned by web-search and does not
%have access to any labelled examples.

\section{Experiments}

We evaluate our method in three settings: (1) Noisy MNIST digit classification, where we add noise and diversity to MNIST~\cite{MNIST} to study our method in a controlled setting, (2) challenging Sports-1M~\cite{Karpathy_CVPR14} action classification for videos in the wild, and (3) newly emerging and fine-grained classes where annotated data is scarce.

\begin{figure*}
  \centering
  \includegraphics[width=0.95\linewidth]{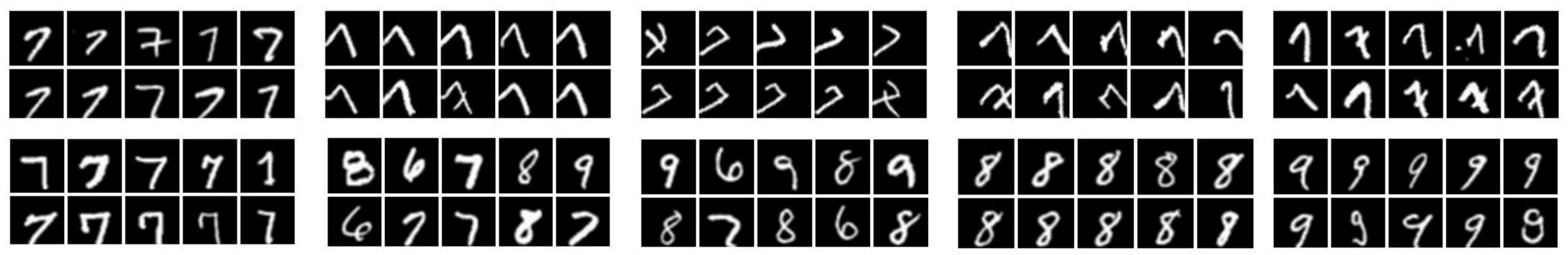}
  \caption{\small{Ten sample query subsets in Noisy MNIST for the digit 7. \emph{Top row.} Different translation and rotation transformations. \emph{Bottom row.} The two leftmost queries have different amounts of noise, the center one is a mixture bucket, and the rightmost two are different digits.}
}
  \label{fig:mnist_learn_set}
  \vspace{-8pt}
\end{figure*}

\begin{figure*}
  \centering
  \includegraphics[width=0.95\linewidth]{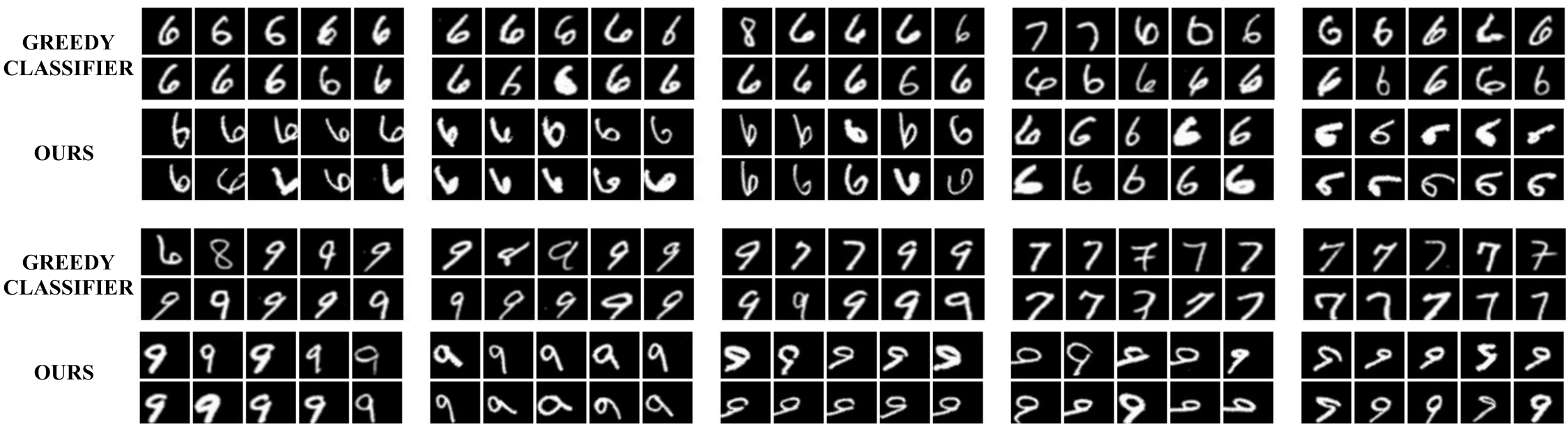}
  \caption{\small{Positive query subsets selected using our method versus the greedy classifier baseline. Subsets chosen by each method are shown from left to right. \emph{Top example.} The greedy classifier chooses queries with very similar content that would not be useful additional positives, whereas our model chooses diverse queries representing different subconcepts of translation and rotation. \emph{Bottom example.} The greedy classifier chooses subsets for the positive digit 9 that include increasingly more noise, and eventually chooses a subset with similar-looking 7s that causes it to begin selecting digit 7 subsets instead. In contrast, our model learns policies that are robust to semantic drift.}}
  \label{fig:mnist_qualitative}
  \vspace{-12pt}
\end{figure*}

\noindent \textbf{Setup.} We evaluate our model on test classes unseen during training. For every test class, the model selects positive examples from $D_{cand}$ and uses them to train a 1-vs-all classifier. We report the mean average precision (mAP) of this classifier on a manually annotated test set. We consider three scenarios, where the maximum number of positive examples chosen from $D_{cand}$ is limited to $60$, $80$ and $100$ respectively. This allows us to measure performance trade-offs at different budgets. Classifiers are initialized with $D_{seed}$ containing the first 10 web-search results.

%In both datasets, we evaluate the quality of the positive examples
%selected from $D_{cand}$ of new unseen test classes. The selected examples are used to train 1-vs-all classifiers for each class. We report the mean average precision (mAP) of the classifiers on a manually labeled set of test class examples. All the data collection models were initialized with the same seed of 10 positive examples, obtained from the top results returned from web-search for a concept.

\noindent \textbf{Baselines.} We compare our model with multiple baselines:
\begin{enumerate} 
\setlength\itemsep{0em}
\vspace{-2pt}

\item \emph{Seed.} Support vector machine (SVM)
learned only from the positive $D_{seed}$.
\vspace{-2pt}
  \item \emph{Label propagation.} Semi-supervised model from
\cite{Zhu_ICML03} used in an inductive setting to learn from the test
$D_{cand}$ and classify the labeled test set.
\vspace{-2pt}

  \item \emph{Label spreading.} Semi-supervised model from
\cite{Zhou_NIPS04} in an inductive setting.
\vspace{-2pt}

  \item \emph{TSVM.} Transductive support vector machine from
\cite{Joachims_ICML99}, which learns a classifier from $D_{cand}$ of test class, and cross-validated using $D_{reward}$. 
\vspace{-2pt}

  \item \emph{Greedy classifier.} An iterative model similar in spirit to~\cite{chen2013neil,li2010optimol} that alternates between greedily selecting queries with the highest-scoring contents according to a classifier, and updating the classifier with the newly labeled examples. We use the same classifier model as in our method. On Sports-1M we compare with two additional variants: \emph{Greedy-Clustering}, which explicitly achieves diversity by clustering labeled positives and ensuring all clusters are represented in new selections; and \emph{Greedy-KL} which balances diversity and drift by selecting queries whose classifier score distribution most closely achieves a ratio of 0.6 (determined through cross-validation) for KL-divergence with the classifier score distribution of $D_{pos}$ vs. of $D_{neg}$.
\vspace{-2pt}

\end{enumerate}

%Each of the baselines along with our model are evaluated in 3 different setups

% Discuss results
\begin{table}
\footnotesize
\centering
% \begin{tabulary}{\linewidth}{|L|C|C|C|C|C|C|}
\begin{tabulary}{\linewidth}{CC|CCCCCC}
\toprule
Digit & Budget & Seed & Label prop. & Label spread. & TSVM & Greedy & Ours\\
\midrule
& 60 & 42.6 & 37.9 & 41.1 & 39.5 & 43.1 & \textbf{60.9}\\
6 & 80 & 42.6 & 40.8 & 45.6 & 44.4 & 43.2 & \textbf{61.3}\\
& 100 & 42.6 & 42.2 & 46.7 & 46.2 & 42.4 & \textbf{71.4}\\
\midrule
& 60 & 48.4 & 51.1 & 48.6 & 46.1 & 49.7 & \textbf{55.1}\\
7 & 80 & 48.4 & 48.8 & 48.5 & 42.6 & 48.5 & \textbf{57.6}\\
& 100 & 48.4 & 48.1 & 46.6 & 39.7 & 47.4 & \textbf{55.7}\\
\midrule
& 60 & 39.1 & 35.0 & 35.2 & 41.2 & 38.3 & \textbf{56.2}\\
8 & 80 & 39.1 & 40.0 & 34.1 & 39.6 & 39.6 & \textbf{55.6}\\
& 100 & 39.1 & 42.0 & 30.2 & 40.8 & 38.0 & \textbf{55.5}\\
\midrule
& 60 & 37.9 & 37.5 & 36.5 & 41.4 & 52.4 & \textbf{52.4}\\
9 & 80 & 37.9 & 37.9 & 37.4 & 38.9 & 53.5 & \textbf{53.5}\\
& 100 & 37.9 & 38.0 & 37.6 & 39.5 & 55.7 & \textbf{55.7}\\
\midrule[1pt]
 & 60 & 42.0 & 40.4 & 40.3 & 42.1 & 43.4 & \textbf{56.1}\\
All & 80 & 42.0 & 41.9 & 41.4 & 41.4 & 43.4 & \textbf{57.0}\\
& 100 & 42.0 & 42.6 & 40.3 & 41.5 & 42.3 & \textbf{59.5}\\
\end{tabulary}
%\vspace{-0.05in}
\caption{\small AP on Noisy MNIST, with budgets of 60, 80 and 100 corresponding to the numbers of positives selected from $D_{cand}$.}
\label{table:mnist_results}
\vspace{-8pt}
\end{table}

\iffalse
\begin{table*}
\footnotesize
\centering
% \begin{tabulary}{\linewidth}{|L|C|C|C|C|C|C|}
\begin{tabulary}{\linewidth}{C|CCCCCC}
\toprule
Budget & Seed & Label propagation & Label spreading & TSVM & Iterative classifier & Ours\\
\midrule
60 & 42.0 & 40.4 & 40.3 & 42.1 & 43.4 & \textbf{56.1}\\
80 & 42.0 & 41.9 & 41.4 & 41.4 & 43.4 & \textbf{57.0}\\
100 & 42.0 & 42.6 & 40.3 & 41.5 & 42.3 & \textbf{59.5}\\
\end{tabulary}
%\vspace{-0.05in}
\caption{\small mAP on MNIST with different budgets for the number of selected positive examples.}
\label{table:mnist_results_map}
\end{table*}
\fi

\noindent \textbf{Implementation Details.}
Our Q-network maps 10-d input histograms in the state representation to a common embedding space of 5 dimensions, with a further hidden layer of 64 units on top. Each episode begins with a seed set of 10 labeled examples, and at training time the network chooses a maximum of 100 total labeled examples needed to train a classifier. The classifier model is a 3-layer multi-layer perceptron with 256 units in each hidden layer, on top of 1000-d ResNet~\cite{He2015} features extracted and then pooled from 10 uniformly sampled frames per video for Sports1M, and on top of raw 784-d pixels for Noisy MNIST. Training consists of 500 episodes for Sports1M, and 200 episodes for Noisy MNIST. The Q-network is trained using experience replay, with mini-batch size 64 and base learning rate $0.01$. A learning update for the agent is taken every 4 iterations, and the target network is softly updated every iteration. A temperature of $\tau=100$ was used in the Q-network.

\subsection{Noisy MNIST digit classification} \label{sec:mnist}

We first evaluate our model in a simulation environment on MNIST digit classification~\cite{MNIST}, where we can introduce noise and subconcept diversity in a controlled manner.  The policies for digit selection are learned from the images of six digits $0-5$ and tested on the other four digits $6-9$.
%We simulate the web search results as described below.

%effect of results returned from web search, where there are multiple related queries which provide relevant examples for a class label.  Our goal is to learn a good policy to select the right examples from such queries in the presence of noise.

\noindent \textbf{Setup.} For every digit class, example images are randomly split into two sets $S_c$ and $S_r$ of $500$ images each. The $D_{cand}$ set is constructed using $S_c$ by simulating both the sub-concept variation and the noise present in web search results. Concretely, the $S_c$ examples are further split into $10$ different query subsets, each containing $5$ sets of $10$
images each. Controlled noise is then added to each query subset: either a specific transformation of the digit (translation and/or rotation), noise from mixed digits, or a different digit.  Examples of query subsets for $D_{cand}$ for the digit $7$ are shown in Fig.~\ref{fig:mnist_learn_set}. The $D_{reward}$ set for a digit is constructed from $S_r$ by applying translation and/or rotation similar to $D_{cand}$, combined with $1000$ negative images sampled from other digits.

%Different amounts of noise are added to each subset.

%We use these query subsets to simulate both sub-concept variation and noise present in web search. Within the $D_{cand}$  set for a digit, each query  Furthermore, the sets within each query contain different amounts of noise. The \emph{reward-set} for a digit consists of $500$ different images of the digit subject to translation and/or rotations of  transformations similar to the candidate set, along with $1000$ negative images sampled from other digits.

\noindent \textbf{Training.} Policies were learned on the
$D_{cand}$ set of the training digits to optimize classifier accuracy on
the annotated $D_{reward}$ examples. Each episode during training requires the model to pick the best $10$ query subsets from $100$ randomly sampled query subsets of the corresponding candidate set. $D_{neg}$ is constructed by randomly sampling $500$ negative examples of other digits. The model parameters were selected by 3-fold cross validation on training digits.

\noindent \textbf{Testing.} The Q-network is used to select a set of $60$, $80$ or $100$
examples from the $D_{cand}$ set of the test
digits. The classifier  is trained with the selected positives and a set of negatives sampled from $D_{cand}$ of other digits. The same $D_{cand}$ examples are used by all baseline methods as well. Classifier accuracy is evaluated on the annotated test set.

\begin{figure*}
  \centering
  {\includegraphics[width=\linewidth]{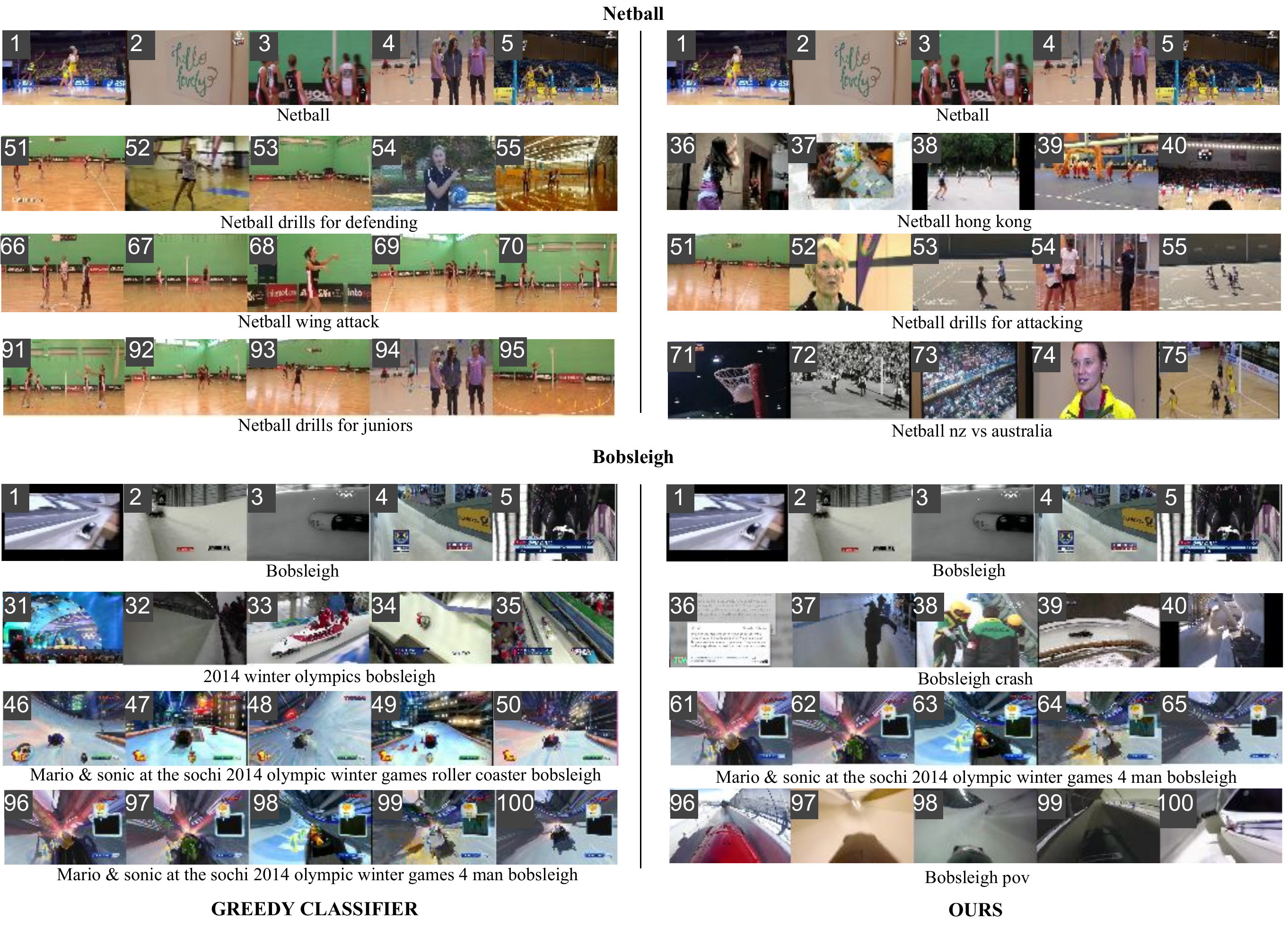}} \\
  \caption{\small{Comparison of positive query subsets selected using our method versus the greedy classifier baseline, for two Sports-1M test classes. Rather than show all 100 selected videos, we highlight interesting differences and show the remaining in Appendix. Each row shows a selected query subset (query phrase and corresponding 5 videos), with the numerical position of the selection out of 100. The first row shows seed videos. \textit{Top example.} The greedy classifier chooses many similar-looking examples, while our method learns that examples of the action in different environments are useful positives. \textit{Bottom example.} The greedy classifier drifts from bobsleigh to video games, while our method is robust to semantic drift and selects useful subcategories of bobsleigh videos such as crashes and pov.}}
  \label{fig:sports1m_qualitative}
  \vspace{-14pt}
\end{figure*}

\noindent \textbf{Results.}
Table~\ref{table:mnist_results} shows quantitative results from our method compared to the baselines, at varying quantities of positive examples chosen by the methods. Our method outperforms all other baselines at all levels by at least $12.7\%$ mAP. Interestingly, as more positive examples are collected (from 60 to 100), baselines typically improve only slightly or even drop in accuracy, whereas our model consistently improves. Traditional semi-supervised methods such as TSVM, label propagation and spreading are designed to select examples similar to the seed set and thus are unable to cope with the large subclass variations. % and hence avoid the transformed variations of the digit.

% For example, after selecting 100 examples, our model is able to train visual classifiers for the test classes with $59.5\%$ mAP; the closest competitor is  $16.9\%$ lower, only achieving $42.6\%$ mAP. 

Qualitative comparison of our method with the strongest greedy classifier baseline is shown in Fig.~\ref{fig:mnist_qualitative}. It illustrates two major pitfalls of greedy methods: (1) in the first example of the digit 6, the greedy classifier selects images overly similar to the seed-set, to the point of including noise, and (2) in the second example of the digit 9, it gets carried away by semantic drift. Our method is  more robust, opting for a diverse selection of subconcepts. It is able to learn that rotations and translations are useful positives for the domain of digits. Interestingly, it tends to gradually expand its understanding of subconcepts, selecting subtle transformations first before more extreme ones. This flexibility to trade-off variety with intra-class similarity allows our model to outperform other methods.

%\subsection{Sports-1M action recognition}
%\label{sec:sports1m}
%\begin{table*}
%\footnotesize
%\centering
% \begin{tabulary}{\linewidth}{|L|C|C|C|C|C|C|}
%\begin{tabulary}{\linewidth}{L|CCCCCC}
%\toprule
%Budget & Seed & Label propagation & Label spreading & TSVM & Greedy classifier & Ours\\
%\midrule
%60 & 64.3 & 65.4 & 65.4 & 70.7 & 71.7 & \textbf{75.4}\\
%80 & 64.3 & 65.4 & 66.6 & 71.7 & 73.8 & \textbf{76.2}\\
%100 & 64.3 & 67.2 & 67.3 & 72.5 & 74.8 & \textbf{77.0}\\
%\end{tabulary}
%\vspace{-0.05in}
%\caption{{\small mAP on sports1M with different budgets for the number of selected positive examples.}}
%\label{table:sports1m_results}
%\end{table*}

\subsection{Sports-1M action recognition}
\label{sec:sports1m}
\begin{table}
\footnotesize
\centering
% \begin{tabulary}{\linewidth}{|L|C|C|C|C|C|C|}
\begin{tabulary}{\linewidth}{L|CCC}
\toprule
Method & Budget-60 & Budget-80 & Budget-100 \\
\midrule
Seed & 64.3 & 64.3 & 64.3 \\
Label propagation & 65.4 & 65.4 & 67.2 \\
Label spreading & 65.4 & 66.6 & 67.3 \\
TSVM & 71.6 & 72.7 & 73.6 \\
Greedy & 71.7 & 73.8 & 74.8 \\
Greedy-clustering & 72.3 & 73.2 & 74.3 \\
Greedy-KL & 74.1 & 74.7 & 74.7 \\
\midrule
\textbf{Ours} & \textbf{75.4} & \textbf{76.2} & \textbf{77.0}
\end{tabulary}
%\vspace{-0.05in}
\caption{{\small mAP on Sports-1M with different budgets for the number of selected positive examples.}}
\label{table:sports1m_results}
\vspace{-12pt}
\end{table}

We evaluate our method in a real-world setting where we want to
classify the $487$ human actions in the Sports-1M video dataset~\cite{Karpathy_CVPR14}.  Collecting high-quality video
examples for human actions can be very laborious and expensive. Hence, we
wish to learn classifiers using only the videos returned by a web
search engine without the need for human annotation.
The classifiers are trained on noisy examples from YouTube and tested on Sports-1M test videos with ground-truth annotations.  We remove any overlapping videos between Sports-1M and
YouTube search results to avoid mixing of training and test data. Throughout this section, we ignore the Sports-1M training videos.

We use $300$ classes for training, $105$ classes for testing and the
remaining classes for validation, and note that this task is fundamentally different from standard 487-way Sports-1M classification. There is no intersection
between the training, testing and validation classes.
%We first explain the construction of \emph{candidate-sets} and \emph{reward-sets} for this setup followed by the training and test setups.

\noindent \textbf{Setup.} The videos returned by YouTube query search were used to construct the candidate sets for both training
and test classes. We constructed $30$ different query expansions for each class using the YouTube query suggestion feature.
The top $30$ videos returned from each query were then split into $6$
different pages of 5 videos each. $20$ queries are sampled for each episode during training, resulting in candidate sets of $600$ videos
per action class split into $120$ different subsets. The annotated Sports-1M videos of the training classes serve as reward sets used to train the Q-network. Sports-1M videos of test classes are used for evaluation.

\noindent \textbf{Training.} In each training episode, the Q-network policies select $20$ subsets ($100$ videos) from the $120$ different query splits of the candidate-set. At each iteration of the episode, the selected positive examples are combined with $500$ random examples from other classes to update the classifier.

\noindent \textbf{Testing and validation.} The model parameters were chosen by cross-validation on the validation classes. The final policies are
used to select positive examples of test classes from the corresponding YouTube search results. These videos are used to train separate 1-vs-all classifiers for each test class. Each classifier is then evaluated on the annotated positive examples of the corresponding class and $1000$ negative examples sampled from videos of other classes.

\begin{figure*}
  \centering
  {\includegraphics[width=\linewidth]{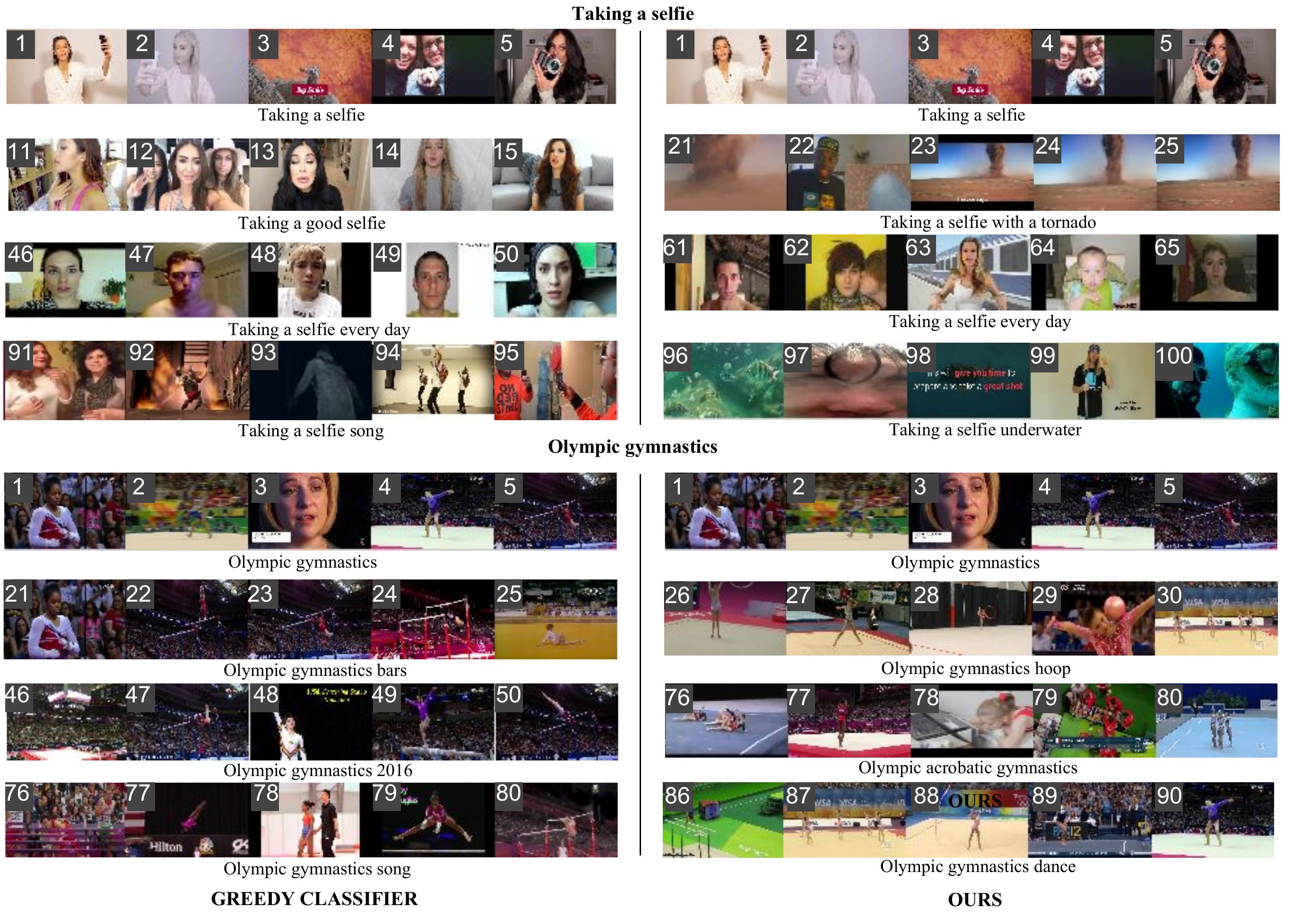}} \\
  \caption{\small{Comparison of positive query subsets selected using our method versus the greedy classifier baseline, for two long-tail classes. See Fig.~\ref{fig:sports1m_qualitative} caption for figure format.
  %All selected 100 videos are shown as thumbnails, with interesting differences highlighted with the corresponding queries. 
  \textit{Top example.} The greedy classifier selects many similar-looking examples of taking a selfie, while our method learns domain-specific knowledge that positives in different environments are more useful, e.g. with a tornado or underwater. \textit{Bottom example.} The greedy classifier selects similar examples of gymnastics, whereas our method selects visually distinct subcategories.}}
  \label{fig:trending_qualitative}
  \vspace{-14pt}
\end{figure*}

\noindent \textbf{Results.}
Table~\ref{table:sports1m_results} compares our model with baselines. Our model outperforms at all budgets, and at Budget-100 by $2.2\%$ mAP. The margin over the Greedy baseline is higher at smaller budgets, indicating that our model more quickly selects the best positive examples. While Greedy-clustering and Greedy-KL improve over Greedy at small budgets, they perform worse at Budget-100. This illustrates that while using heuristics to explicitly balance diversity and drift can help early on, it is hard to ultimately avoid noise.

% The margins increase with smaller budgets, indicating that our model quickly selects the best positive examples. For example, after selecting 60 examples, our model is able to train visual classifiers for the test classes which achieve $3.7\%$ higher mAP than the closest competitor (the greedy classifier). 

Fig.~\ref{fig:sports1m_qualitative} shows positive examples selected by our method compared to the greedy classifier. In the top example of netball, the greedy classifier is overly conservative and selects positives very similar to existing positives, which does not improve classifier accuracy. On the other hand, our model learns domain-specific knowledge that examples of the target action in visually different environments are useful. In the bottom example of bobsleigh, the greedy classifier suffers from the opposite problem: after the classifier selects some queries containing video games, it drifts to queries with more and more video games. In contrast, our model is more robust and returns to clean bobsleigh videos with minimal drift. Furthermore, it selects different subcategories of bobsleigh videos: crash videos, and pov videos, which are useful for training a classifier.

%These two examples demonstrate that our model is able to effectively capture visual diversity while minimizing semantic drift.

% \begin{figure*}
%   \centering
%   {\includegraphics[width=\linewidth]{fig/taking_a_selfie_final.pdf}} \\
%   \rule{\linewidth}{1pt} \\
%   {\includegraphics[width=\linewidth]{fig/olympic_gymnastics_final.pdf}}
%   \caption{\small{Comparison of positive examples selected using our method versus the iterative classifier similar to~\cite{chen2013neil,li2010optimol}. 
%  Rather than show all 100 selected videos, we highlight the interesting differences \TODO{...}
%   %All selected 100 videos are shown as thumbnails, with interesting differences highlighted with the corresponding queries. 
%   \textit{Top example.} The iterative classifier chooses many similar-looking examples of selfie, while our method chooses selfies in different environments such as with a tornado and underwater. \textit{Bottom example.} The iterative classifier chooses similar examples of gymnastics, whereas our method chooses subcategories of gymnastics.} \TODO{OLGA: yikes, i don't understand this figure at all. Remove thumbnails, those are not useful. Make the action label clearly visible above the videos. Remove the quotation marks. Iterative classifier on the left, ours on the right. Make it clear there are two examples, otherwise it looks like 6}}
%   \label{fig:trending_qualitative}
% \end{figure*}

\subsection{Long-tail action labeling} \label{sec:trending}

We show examples of the policy we learned for Sports-1M on new action classes for which annotated data does not exist in Fig.~\ref{fig:trending_qualitative}. We compare our learned policy vs. the greedy classifier, for a recent societal-concept: ``Taking a selfie'', and a fine-grained class: ``Olympic gymnastics''.

%We followed the methodology of Sec.~\ref{sec:sports1m}.

%to generate 20 candidate YouTube queries from autocomplete search results, and scraped 30 videos/query. 

%Additional examples are shown in Supplementary.

The videos selected in Fig.~\ref{fig:trending_qualitative} for ``Taking a selfie'' show that the policy again utilizes domain-specific knowledge that an action in different environments is useful for positive examples: e.g. in front of a volcano, and underwater, even though these have diverse visual appearance. In contrast, the greedy classifier tends to select videos that look very similar to the seed videos.  The videos selected for ``Olympic gymnastics'' demonstrate that the policy selects visually different subcategories: gymnastics with hoops, and gymnastics with acrobatics.

Table~\ref{table:trending_stats} measures the diversity and correctness of our model versus the greedy model. \emph{Query recall} is the number of correct queries which contributed to the selected positives. The correct queries were manually annotated. Our model selects positives from more queries promoting higher diversity. Similarly, our model also avoids noisy examples as seen from \emph{video recall}: the number of true-positive videos included in the 100 videos selected by each model.

%The true-positives in the candidate set were manually annotated.

%shows statistics of the selected videos after de-duping across queries, where query recall is the number of manually-annotated true positive subconcepts (defined as distinct queries) that are selected, and video recall is the percentage of unique true positive videos that are selected compared to an oracle (where all 100 selections are unique true positives).  In both examples, our learned policy has both higher query recall and video recall, showing that it successfully selects for visually diverse examples while avoiding semantic drift.

\begin{table}[bt]
\begin{center}
\footnotesize
\begin{tabulary}{0.5\linewidth}{c c c | c c}
& \multicolumn{2}{c}{Query recall} & \multicolumn{2}{c}{Video recall} \\
\midrule
Class & Greedy & Ours & Greedy & Ours\\
\midrule
Taking a selfie & 6/16 & \textbf{9/16} & 75\% & \textbf{90\%} \\
Olympic gymnastics & 7/18 & \textbf{10/18} & 76\% & \textbf{82\%}\\
\end{tabulary}
\vspace{-8pt}
\end{center}
\caption{\small{Query and video recall of positive videos for two long-tail classes, for our method vs. the greedy classifier. Our method has higher recall of true-positive queries and videos, showing that it selects diverse subconcepts while avoiding semantic drift.}}
\label{table:trending_stats}
\vspace{-16pt}
\end{table}

% \begin{table}[bt]
% \begin{center}
% \footnotesize
% \begin{tabulary}{0.5\linewidth}{c c c c c}

% & \multicolumn{2}{Query recall} &\multicolumn{2}{Video recall} \\
% Class & Iterative & Ours & Iterative & Ours
% \midrule
% Taking Selfie - Iterative & 6/16 & 75\%\\
% Selfie - Ours & \textbf{9/16} & \textbf{90\%}\\
% \midrule
% Oly. gymnastics - Iterative & 7/18 & 76\%\\
% Oly. gymnastics - Ours & \textbf{10/18} & \textbf{82\%}\\
% \end{tabulary}
% \end{center}
% \caption{\small{Statistics of positive videos labeled by our policy for the long-tail ``Taking a selfie" and ``Olympic gymnastics" action classes, for our method vs. the iterative classifier baseline. Across both classes, our method has higher recall of manually-determined true-positive queries and videos, and greater total number of unique positive vieos selected, showing that it selects a greater diversity of subconcepts.} \TODO{Olga: I could help if I understood what this table is showing. Happy to talk if that would be helpful...}}
% \label{table:trending_stats}
% \end{table}
\section{Conclusion}
\label{sec:conclusion}
In conclusion, we have introduced a principled, reinforcement learning-based formulation for learning how to label noisy web data. We show that our method is able to learn domain-specific knowledge, and label data for new classes in a way that achieves diversity while avoiding semantic drift. We demonstrate our method first in the controlled setting of MNIST, then on large-scale Sports-1M, and finally on newly emerging and fine-grained classes.
\section*{Acknowledgments}
\label{sec:acknowledgments}
Our work is supported by an ONR MURI grant and a hardware donation from NVIDIA.

{\small
\bibliographystyle{ieee}
\bibliography{video_selection_bib}

\begin{thebibliography}{10}\itemsep=-1pt

\bibitem{Andrews_NIPS02}
S.~Andrews, I.~Tsochantaridis, and T.~Hofmann.
\newblock Svms for multiple-instance learning.
\newblock In {\em NIPS}, 2002.

\bibitem{icoseg}
D.~Batra, A.~Kowdle, D.~Parikh, J.~Luo, and T.~Chen.
\newblock {iCoseg}: Interactive co-segmentation with intelligent scribble
  guidance.
\newblock In {\em CVPR}, 2010.

\bibitem{brazdil2008metalearning}
P.~Brazdil, C.~G. Carrier, C.~Soares, and R.~Vilalta.
\newblock {\em Metalearning: applications to data mining}.
\newblock Springer Science \& Business Media, 2008.

\bibitem{carlson2010toward}
A.~Carlson, J.~Betteridge, B.~Kisiel, B.~Settles, E.~R. Hruschka~Jr, and T.~M.
  Mitchell.
\newblock Toward an architecture for never-ending language learning.
\newblock In {\em AAAI}, volume~5, page~3, 2010.

\bibitem{Chen15}
X.~Chen and A.~Gupta.
\newblock Webly supervised learning of convolutional networks.
\newblock In {\em ICCV}, 2015.

\bibitem{chen2015webly}
X.~Chen and A.~Gupta.
\newblock Webly supervised learning of convolutional networks.
\newblock In {\em Proceedings of the IEEE International Conference on Computer
  Vision}, pages 1431--1439, 2015.

\bibitem{chen2013neil}
X.~Chen, A.~Shrivastava, and A.~Gupta.
\newblock {NEIL}: Extracting visual knowledge from web data.
\newblock In {\em ICCV}, 2013.

\bibitem{LEVAN}
S.~K. Divvala, A.~Farhadi, and C.~Guestrin.
\newblock Learning everything about anything: Webly-supervised visual concept
  learning.
\newblock In {\em CVPR}, 2014.

\bibitem{dulac2015reinforcement}
G.~Dulac-Arnold, R.~Evans, P.~Sunehag, and B.~Coppin.
\newblock Reinforcement learning in large discrete action spaces.
\newblock {\em arXiv preprint arXiv:1512.07679}, 2015.

\bibitem{feurer2015efficient}
M.~Feurer, A.~Klein, K.~Eggensperger, J.~Springenberg, M.~Blum, and F.~Hutter.
\newblock Efficient and robust automated machine learning.
\newblock In {\em NIPS}, 2015.

\bibitem{frome2013devise}
A.~Frome, G.~Corrado, et~al.
\newblock Devise: A deep visual-semantic embedding model.
\newblock In {\em NIPS}, 2013.

\bibitem{gan2016webly}
C.~Gan, C.~Sun, L.~Duan, and B.~Gong.
\newblock Webly-supervised video recognition by mutually voting for relevant
  web images and web video frames.
\newblock In {\em European Conference on Computer Vision}, pages 849--866.
  Springer, 2016.

\bibitem{He2015}
K.~He, X.~Zhang, S.~Ren, and J.~Sun.
\newblock Deep residual learning for image recognition.
\newblock {\em CoRR:1512.03385}, 2015.

\bibitem{Joachims_ICML99}
T.~Joachims.
\newblock Transductive inference for text classification using support vector
  machines.
\newblock In {\em ICML}, 1999.

\bibitem{joulin2010discriminative}
A.~Joulin, F.~Bach, and J.~Ponce.
\newblock Discriminative clustering for image co-segmentation.
\newblock In {\em CVPR}, 2010.

\bibitem{Karpathy_CVPR14}
A.~Karpathy, G.~Toderici, S.~Shetty, T.~Leung, R.~Sukthankar, and L.~Fei-Fei.
\newblock Large-scale video classification with convolutional neural networks.
\newblock In {\em CVPR}, 2014.

\bibitem{Kingma_NIPS14}
D.~Kingma, D.~Rezende, S.~Mohamed, and M.~Welling.
\newblock Semi-supervised learning with deep generative models.
\newblock In {\em NIPS}, 2014.

\bibitem{MNIST}
Y.~LeCun, L.~Bottou, Y.~Bengio, and P.~Haffner.
\newblock Gradient-based learning applied to document recognition.
\newblock {\em Proceedings of the IEEE}, 86:2278--2324, 1998.

\bibitem{li2010optimol}
L.-J. Li and L.~Fei-Fei.
\newblock {OPTIMOL}: automatic online picture collection via incremental model
  learning.
\newblock {\em IJCV}, 88(2):147--168, 2010.

\bibitem{li2015semi}
X.~Li, Y.~Guo, and D.~Schuurmans.
\newblock Semi-supervised zero-shot classification with label representation
  learning.
\newblock In {\em Proceedings of the IEEE International Conference on Computer
  Vision}, pages 4211--4219, 2015.

\bibitem{liang2016learning}
J.~Liang, L.~Jiang, D.~Meng, and A.~Hauptmann.
\newblock Learning to detect concepts from webly-labeled video data.
\newblock In {\em Joint Conference on Artificial Intelligence (IJCAI)}, 2016.

\bibitem{mnih2015human}
V.~Mnih, K.~Kavukcuoglu, D.~Silver, A.~A. Rusu, J.~Veness, M.~G. Bellemare,
  et~al.
\newblock Human-level control through deep reinforcement learning.
\newblock {\em Nature}, 518(7540):529--533, 2015.

\bibitem{palatucci2009zero}
M.~Palatucci, D.~Pomerleau, G.~E. Hinton, and T.~M. Mitchell.
\newblock Zero-shot learning with semantic output codes.
\newblock In {\em Advances in neural information processing systems}, pages
  1410--1418, 2009.

\bibitem{Pitelis_KDD14}
N.~Pitelis, C.~Russell, and L.~Agapito.
\newblock Semi-supervised learning using an unsupervised atlas.
\newblock In {\em Machine Learning and Knowledge Discovery in Databases}, pages
  565--580. Springer, 2014.

\bibitem{ranzato2008semi}
M.~Ranzato and M.~Szummer.
\newblock Semi-supervised learning of compact document representations with
  deep networks.
\newblock In {\em ICML}, 2008.

\bibitem{Rifai_NIPS11}
S.~Rifai, Y.~N. Dauphin, P.~Vincent, Y.~Bengio, and X.~Muller.
\newblock The manifold tangent classifier.
\newblock In {\em NIPS}, 2011.

\bibitem{rohrbach2011evaluating}
M.~Rohrbach, M.~Stark, and B.~Schiele.
\newblock Evaluating knowledge transfer and zero-shot learning in a large-scale
  setting.
\newblock In {\em CVPR}, 2011.

\bibitem{schroff2011harvesting}
F.~Schroff, A.~Criminisi, and A.~Zisserman.
\newblock Harvesting image databases from the web.
\newblock {\em TPAMI}, 2011.

\bibitem{watkins1992q}
C.~J. Watkins and P.~Dayan.
\newblock Q-learning.
\newblock {\em Machine learning}, 8(3-4):279--292, 1992.

\bibitem{weston2012deep}
J.~Weston, F.~Ratle, H.~Mobahi, and R.~Collobert.
\newblock Deep learning via semi-supervised embedding.
\newblock In {\em Neural Networks: Tricks of the Trade}, pages 639--655.
  Springer, 2012.

\bibitem{xu2016multi}
X.~Xu, T.~M. Hospedales, and S.~Gong.
\newblock Multi-task zero-shot action recognition with prioritised data
  augmentation.
\newblock In {\em European Conference on Computer Vision}, pages 343--359.
  Springer, 2016.

\bibitem{zaremba2015learning}
W.~Zaremba, T.~Mikolov, A.~Joulin, and R.~Fergus.
\newblock Learning simple algorithms from examples.
\newblock {\em CoRR:1511.07275}, 2015.

\bibitem{Zhou_NIPS04}
D.~Zhou, O.~Bousquet, T.~N. Lal, J.~Weston, and B.~Sch{\"o}lkopf.
\newblock Learning with local and global consistency.
\newblock {\em NIPS}, 2004.

\bibitem{Zhu_ICML03}
X.~Zhu and Z.~Ghahramani.
\newblock Learning from labeled and unlabeled data with label propagation.
\newblock In {\em ICML}, 2002.

\end{thebibliography}
}

\end{document}